\documentclass[a4paper,fleqn]{cas-dc}

\usepackage[utf8]{inputenc}
\usepackage{csquotes}  
\usepackage[T1]{fontenc}    
\usepackage{hyperref}       
\usepackage{url}            
\usepackage{booktabs}       
\usepackage{amsfonts}       
\usepackage{nicefrac}       
\usepackage{microtype}      
\usepackage{lipsum}
\usepackage{fancyhdr}       
\usepackage{graphicx}       
\usepackage{multirow}
\usepackage{tabularx}
\usepackage{array} 
\usepackage{caption}
\usepackage{amsmath}
\usepackage{booktabs,tabularx, multirow, array}

\graphicspath{{media/}}     

\usepackage[english]{babel}

\usepackage[style=apa, backend=biber]{biblatex}
\DeclareLanguageMapping{english}{english-apa}
\usepackage{soul}

\def\tsc#1{\csdef{#1}{\textsc{\lowercase{#1}}\xspace}}
\tsc{WGM}
\tsc{QE}

\begin{document}
\let\WriteBookmarks\relax
\def\floatpagepagefraction{1}
\def\textpagefraction{.001}

\shorttitle{DeBERTa-KC: A Knowledge Construction Classifier}    

\shortauthors{}  

\title [mode = title]{Analysing Knowledge Construction in Online Learning: Adapting the Interaction Analysis Model for Unstructured Large-Scale Discourse}  

\begin{abstract}
The rapid expansion of online courses and social media has generated large volumes of unstructured learner-generated text. Understanding how learners construct knowledge in these spaces is crucial for analysing learning processes, informing content design, and providing feedback at scale. However, existing approaches typically rely on manual coding of well-structured discussion forums, which does not scale to the fragmented discourse found in online learning. This study proposes and validates a framework that combines a codebook inspired by the Interaction Analysis Model with an automated classifier to enable large-scale analysis of knowledge construction in unstructured online discourse. We adapt four comment-level categories of knowledge construction: Non-Knowledge Construction, Share, Explore, and Integrate. Three trained annotators coded a balanced sample of 20{,}000 comments from YouTube education channels. The codebook demonstrated strong reliability, with Cohen's $\kappa = 0.79$ on the main dataset and $0.85$--$0.93$ across four additional educational domains. For automated classification, bag-of-words baselines were compared with transformer-based language models using 10-fold cross-validation. A DeBERTa-v3-large model achieved the highest macro-averaged F1 score ($0.841$), outperforming all baselines and other transformer models. External validation on four domains yielded macro-F1 above $0.705$, with stronger transfer in medicine and programming, where discourse was more structured and task-focused, and weaker transfer in language and music, where comments were more varied and context-dependent. Overall, the study shows that theory-driven, semi-automated analysis of knowledge construction at scale is feasible, enabling the integration of knowledge-construction indicators into learning analytics and the design of online learning environments.

\end{abstract}



\author[1]{Jindi Wang}
\author[2]{Yidi Zhang}
\author[3]{Zhaoxing Li*}
\author[2]{Pedro Bem-Haja}
\author[1]{Ioannis Ivrissimtzis}
\author[4]{Zichen Zhao}
\author[3]{Sebastian Stein}


\affiliation[1]{%
  organization={Department of Computer Science, Durham University},
  city={Durham},
  country={UK}
}

\affiliation[2]{%
  organization={Department of Education and Psychology, University of Aveiro},
  city={Aveiro},
  country={Portugal}
}

\affiliation[3]{%
  organization={School of Electronics and Computer Science, University of Southampton},
  city={Southampton},
  country={UK}
}

\affiliation[4]{%
  organization={School of Mathematics and Statistics, Lanzhou University},
  city={Lanzhou},
  country={China}
}



\maketitle

\section{Introduction}

Identifying and supporting knowledge construction in online discussions remains a significant challenge in education \parencite{Gunawardena2023}. As online learning environments such as massive open online courses, discussion forums, and social media continue to expand, learners are producing large amounts of unstructured textual data \parencite{Dubovi2020,Nguyen2023}. Understanding how knowledge is constructed within these environments provides an important foundation for analysing the cognitive processes of learners \parencite{Gunawardena1997,Henri1992}. For example, such insights can inform the design of more targeted instructional materials and scaffolds that promote higher-level epistemic moves \parencite{Weinberger2006,Scardamalia2006}, support learning analytics indicators that monitor the quality of discourse over time rather than only counting participation \parencite{gavsevic2015let,wise2013learning}, and help instructors provide more timely and focused feedback by identifying when learners are merely sharing information versus exploring or integrating ideas \parencite{DeWever2006,Dubovi2020}.

Traditional analyses of knowledge construction have relied on manual content analysis. For example, \textcite{Henri1992} proposed a five-dimensional framework for computer-mediated communication that includes participation, interaction, social, cognitive, and metacognitive dimensions, providing a systematic basis for interpreting learning processes in message transcripts. \textcite{Gunawardena1997} introduced the Interaction Analysis Model (IAM), which conceptualises the process of knowledge construction in computer-mediated communication as a progressive sequence of phases, from sharing and comparing information to negotiating meaning, constructing new knowledge, and reaching mutual understanding. The IAM has since become a foundational framework for examining collaborative learning and knowledge building in online environments \parencite{Lucas2014}. Subsequent studies further refined the analysis of knowledge construction. \textcite{Hmelo-Silver2003} employed mixed methods to investigate how groups build shared understanding, demonstrating that detailed conversational exchanges and broader thematic patterns characterise computer-supported collaborative learning. \textcite{Weinberger2006} proposed a multidimensional framework for argumentative knowledge construction, distinguishing participation, epistemic, argumentative, and social dimensions to capture how learners co-construct knowledge through discourse. Together, these studies have advanced methodological approaches for examining knowledge construction in digital learning contexts.

Despite recent progress, manual coding remains labour-intensive, time-consuming, and susceptible to subjectivity \parencite{Dornauer2024}. Studies must establish and maintain inter-rater reliability, and disagreements often require negotiated coding \parencite{Hallgren2012}. To reduce delay and improve scalability, researchers have explored automated discourse classification for timely feedback. Deep learning and other supervised approaches have been used to identify phases of knowledge construction \parencite{Hu2021}. However, performance varies with language, context, and the availability of labelled data, and key challenges persist \parencite{Hu2022}.

Recent advances in large pre-trained language models have substantially improved contextual representation in educational discourse analysis, including work grounded in the Interaction Analysis Model (IAM) \parencite{Castellanos-Reyes2025,sakeef2025detecting,Gunawardena2023}. Transformer architectures such as BERT, RoBERTa, and DeBERTa have demonstrated strong performance on related discourse tasks (e.g., stance detection, sentiment analysis, and argumentation modelling) \parencite{Devlin2019,He2021}. However, their application to fine-grained classification of IAM-based knowledge construction remains limited. Prior studies typically rely on small or imbalanced datasets, employ coarse category schemes, and focus on single, well-structured discussion contexts, with little evaluation of cross-domain robustness \parencite{Dubovi2020,Nguyen2023,Gunawardena2023}.

In addition, models optimised for specific datasets or domains are often quickly outperformed by newer language-model variants, improving raw performance but hindering the development of stable, theory-grounded approaches that can be reused across studies \parencite{Castellanos-Reyes2025,mim2025words}. Consequently, existing automated methods provide only partial support for reliably distinguishing nuanced IAM phases—such as sharing, exploration, and integration—within large-scale, noisy social media discourse \parencite{Nguyen2023,Dubovi2020}.


Therefore, this study aims to propose and validate a framework that integrates an IAM-inspired codebook with an automated classifier trained on a balanced category training set, enabling large-scale analysis of knowledge construction in online learning environments.

The remainder of this section provides the conceptual background for the study. Section~\ref{sec: Identifying knowledge construction} discusses how knowledge construction manifests in online learning discourse, Section~\ref{sec: analyse knowledge construction} reviews the Interaction Analysis Model and our rationale for adapting it to comment-level analysis, and Section~\ref{sec:model limitations} surveys work on automated classification of knowledge construction and related constructs. Section~\ref{sec:aim} then states the overall aim of the study and formulates the research questions. Section~\ref{sec:methods} describes the data, annotation procedure, and modelling pipeline; Section~\ref{sec:results} presents the experimental results; Section~\ref{sec:discussion} discusses implications and limitations; and Section~\ref{sec:conclusion} concludes.

\subsection{Identifying knowledge construction in online learning discourse} \label{sec: Identifying knowledge construction}
Knowledge construction is a cognitive process in which learners actively interpret, organise, and integrate information to build new understanding \parencite{Scardamalia2006}. In online learning environments, this process becomes visible through learners’ discourse and comment interactions, where they articulate claims, question ideas, provide justifications, and integrate multiple perspectives 
\parencite{Nguyen2023,Mayer1996}. Such forms of engagement further encourage learners to set their own goals and regulate their learning \parencite{Greenhow2016, Tazhenova2024}

However, discourse and comment interaction in computer-mediated communication is not always linear. Messages may appear as independent contributions rather than as parts of a structured exchange \parencite{Marcoccia2004}. In many online environments, discussions are asynchronous and lack reciprocal engagement, making it difficult to maintain coherent conversational threads \parencite{Krykoniuk2025}. Participants may misinterpret the conversational structure and it is often unclear who is responding to whom \parencite{Marcoccia2004}. This pattern is even more evident on social media platforms, such as YouTube, where most comments are short, fragmented, and do not form extended exchanges \parencite{Tanskanen2021}.

Despite this structural fragmentation, individual comments often still contain linguistically identifiable elements that correspond to the different phases of knowledge construction such as triggering events, exploration, integration, and resolution, even though these phases do not appear in a strict sequential order \parencite{Garrison2001}. Research in epistemic cognition and argumentation further demonstrates that such discursive moves can reliably indicate forms of knowledge construction even when direct reciprocal interaction is absent \parencite{Nguyen2023, Dubovi2020}. Thus, although asynchronous comment streams lack the structural coherence needed for phase-by-phase interactional analysis, they nevertheless provide analytically meaningful traces of knowledge construction processes \parencite{Gunawardena2023, Ba2023}.

Moreover, although individual comments often appear structurally independent, they may still respond to video content or to the contributions of other viewers \parencite{Krykoniuk2025}. Even without explicit reply chains, comment sections can display coherence and cross-modal cohesion \parencite{Tanskanen2021}. From this perspective, online comment spaces can be understood as multi-threaded dialogic networks in which individual contributions are part of a broader shared discourse and a collective cognitive space \parencite{Kimmerle2015, Ye2022}.

Therefore, although knowledge construction can be identified within individual comments, the loose and fragmented structure of social media content often makes such identification inconsistent. This indicates the need for more systematic and scalable approaches to improve the reliability of identifying knowledge construction in online learning contexts.

\subsection{Using the IAM to analyse knowledge construction} \label{sec: analyse knowledge construction}

The IAM is widely used to examine knowledge construction in online learning \parencite{Gunawardena1997}. It delineates five phases through which learners progressively build knowledge: 1) sharing and comparing information, 2) discovering and exploring dissonance or inconsistency among ideas, concepts or statements, 3) negotiating meaning/co-constructing knowledge, 4) testing and modifying a proposed synthesis or co-construction, and 5) agreeing with statement(s)/applying newly constructed meaning. Although the IAM was originally conceptualised as a structured and progressive interactional process \parencite{Lucas2014}, the IAM’s phases can also be interpreted as epistemic functions that may surface within individual contributions \parencite{Gunawardena2023}.This reinterpretation is particularly relevant in online environments where discourse is often non-linear and fragmented \parencite{Krykoniuk2025}.

Applying the IAM to single comments therefore entails a conceptual shift. Comment-level coding captures the epistemic functions enacted in an individual contribution rather than the interactional progression theorised by \textcite{Gunawardena1997}. This distinction is central to construct validity, while the adapted categories provide evidence of cognitive engagement, they no longer represent the temporal or collaborative trajectory of co-construction \parencite{Nguyen2023}. This interpretation aligns with a broader methodological shift in online discourse analysis, where epistemic moves are examined as linguistic indicators that need not follow a linear sequence \parencite{Gunawardena2023}. From this perspective, even fragmented or weakly threaded digital environments contain analytically meaningful traces of cognitive work, despite the absence of observable interactional progression \parencite{Dubovi2020}.

This approach affords both strengths and limitations. Conceptually, treating discourse as evidence of how learners publicly display their thinking connects comment-level analysis to interaction-oriented models \parencite{Hmelo-Silver2003, Garrison2001}. For example, \textcite{Dubovi2020} observed that comments on YouTube science videos often progressed from simple information exchange to argumentative negotiation, which suggests the potential for deeper knowledge construction. In a similar way, \textcite{Nguyen2023} adapted the IAM to remove its dependence on linear reply structures and found that in TikTok educational content, comments were predominantly social, with knowledge construction emerging mainly through opinion sharing and the exploration of disagreement.

At the same time, reducing the unit of analysis to single comments inevitably constrains the granularity of the model \parencite{Gunawardena1997}. Higher-level processes such as negotiating meaning or co-construction are harder to infer without interactional context \parencite{Lucas2014}, and empirical work shows that these phases occur infrequently even in structured online discussions \parencite{Gunawardena2023}. Moreover, increases in interaction frequency do not necessarily lead to higher levels of knowledge-construction phases \parencite{hew_cheung_2011}. Consequently, work in non-interactive environments has often collapsed later IAM phases into broader, lower-level categories to improve coding reliability \parencite{Nguyen2023}. This pragmatic adjustment does not diminish the value of the IAM; rather, it situates comment-level analyses within an emerging paradigm in which IAM phases function as discourse markers rather than as strict interactional stages \parencite{Dubovi2020}.

Therefore, although the IAM can be effectively applied at the comment level to detect traces of knowledge construction, such use requires careful reconceptualisation of the phases. Ideally, it should also be supported by complementary analytical techniques that strengthen the interpretive validity of phase assignments.

\subsection{Automated classification of knowledge construction}\label{sec:model limitations}
Recent advances in natural language processing have enabled the automated classification of knowledge construction in online learning discourse \parencite{Gunawardena2023}. Early work demonstrated that neural models can identify discourse segments associated with different phases of cognitive presence, establishing the feasibility of such approaches \parencite{Hu2021}. More recently, transformer-based architectures such as BERT \parencite{Devlin2019} and DeBERTa \parencite{He2021} have achieved substantial improvements in fine-grained discourse classification, showing strong agreement with human annotations in educational contexts \parencite{Ba2023}. Extensions incorporating multi-label classification have further improved robustness by capturing the overlapping and fluid nature of epistemic phases \parencite{Hu2023}. Studies integrating behavioural or semantic metadata indicate, however, that textual features remain the primary predictive signal \parencite{Dornauer2024}. At the same time, recent reviews highlight the need to balance performance gains from large language models with interpretability and cross-context robustness \parencite{Castellanos-Reyes2025}.

Despite these advances, automated IAM-based classification faces persistent structural and methodological limitations. The nonlinear and weakly threaded nature of many online platforms—such as YouTube, Facebook, and Twitter/X—obscures conversational context and constrains interaction-level analysis \parencite{Herring1999,Aragon2017,Naab2025,Dubovi2020}. Beyond structural issues, construct validity remains a central concern: machine-learning models infer IAM categories from statistical regularities in text, even though these categories represent cognitive functions that are not directly observable. This creates a risk of reliance on superficial lexical cues, particularly when datasets are small, imbalanced, or domain specific \parencite{Breivik2016,Kovanovic2016,Hu2021,Neto2021,Dornauer2024}. Ensuring validity therefore requires diverse training data, safeguards against overfitting, and evaluation strategies that test alignment with theoretical expectations rather than surface linguistic patterns \parencite{Ferreira2020,Ba2023}.

To accommodate fragmented discourse structures, recent studies have applied IAM coding at the level of individual comments. \textcite{Gunawardena2023} reported modest performance on a small corpus of discussion posts, while \textcite{Nguyen2023} achieved higher F1 scores using BERT on TikTok comments but relied on uneven label distributions and a single-domain setting. Overall, existing work remains limited in scale and scope, with little evidence for cross-domain generalisation. Consequently, larger and more balanced datasets, combined with transformer-based models and external validation across educational domains, are needed to establish reliable and reusable approaches for automated knowledge-construction analysis.

\subsection{Study aim and contributions}\label{sec:aim}

Against this background, this study operationalises the Interaction Analysis Model (IAM) for large-scale, non-structured online comment streams, where contributions are typically short, non-sequential, and weakly threaded (e.g., YouTube comments). We combine a manually developed, IAM-inspired codebook with supervised classification: a balanced, human-annotated sample provides ground truth for training and evaluating automated classifiers that can then be applied at scale to unlabelled comments.

We make three contributions. First, we adapt IAM to a compact four-category, comment-level scheme (\textit{NonKC}, \textit{Share}, \textit{Explore}, \textit{Integrate}) tailored to non-interactive, unstructured discourse and demonstrate its reliability across domains. Second, we construct and (on acceptance) release a 20{,}000-comment, manually annotated corpus and use it to benchmark classical and transformer-based classifiers, showing the practical trade-offs between accuracy and stability. Third, we evaluate cross-domain generalisation by validating the best model on four additional educational domains (language, medicine, music, programming) and analysing per-category transfer behaviour.

These aims are operationalised in three research questions:

\begin{enumerate}
    \item[RQ1:] How can the IAM be adapted into a reliable comment-level coding scheme for non-structured, nonlinear online comments?
    \item[RQ2:] On a large, manually annotated corpus, how do traditional machine-learning and transformer-based models compare in identifying IAM categories, and which configuration balances accuracy and stability?
    \item[RQ3:] To what extent does the best-performing classifier (DeBERTa-KC) generalise to other educational domains?
\end{enumerate}

\section{Methods}
\label{sec:methods}

\begin{figure*}[htbp]
    \centering
    \includegraphics[width=0.9\linewidth]{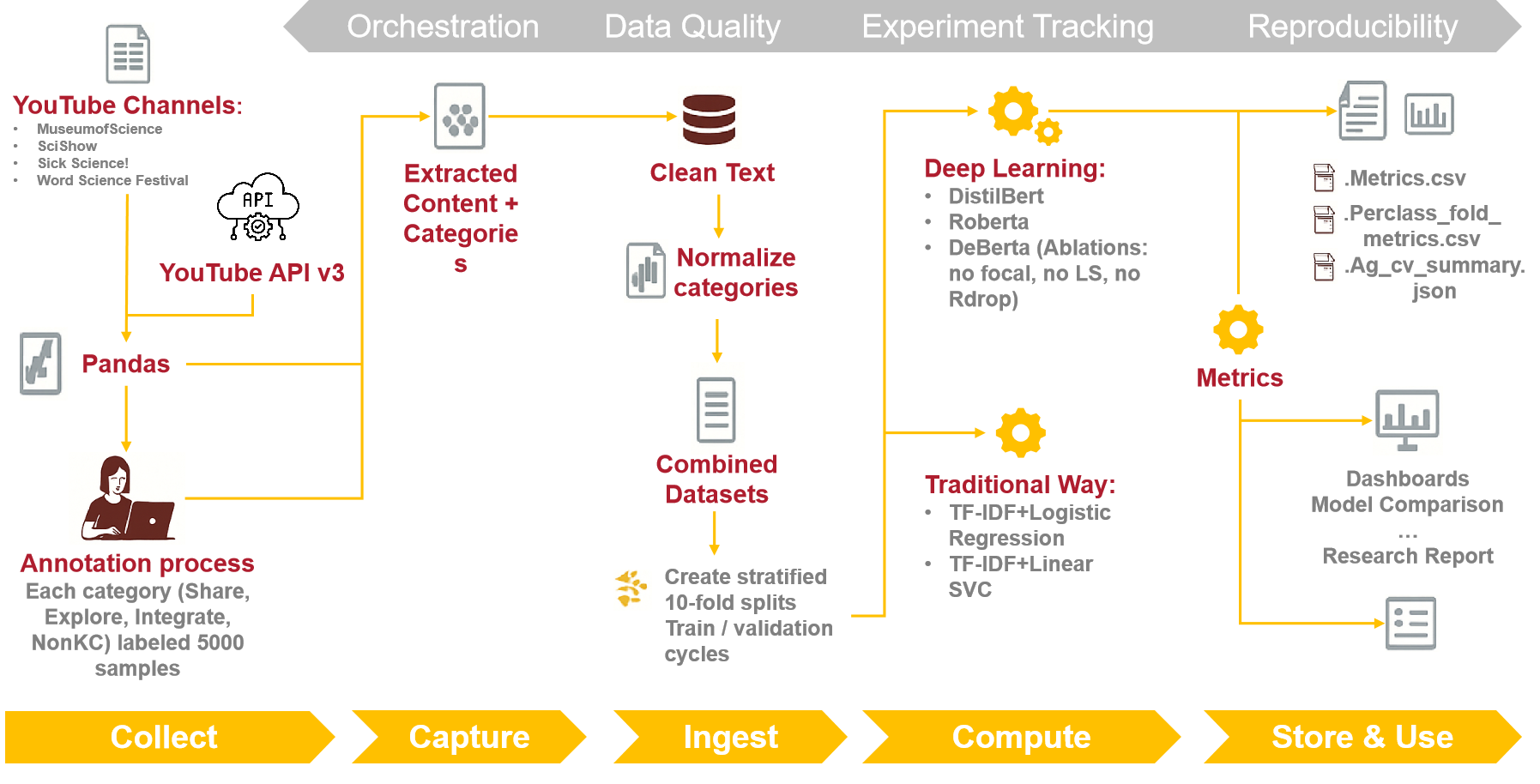}
 \caption{Data collection and model evaluation pipeline.}
    \label{fig:pipeline}
\end{figure*}

Figure~\ref{fig:pipeline} presents an overview of our end-to-end workflow, spanning data collection through model evaluation and reproducibility. The pipeline is organised into five stages: \textit{Collect}, \textit{Capture}, \textit{Ingest}, \textit{Compute}, and \textit{Store \& Use}. It is implemented through modular Python scripts with integrated experiment tracking, and the full source code is available in an anonymised GitHub repository. The repository link will be released upon acceptance to preserve double-blind review. All key decisions in this pipeline (tokenisation, fold splits, hyperparameters, and metrics) are systematically versioned to ensure auditability and to enable exact reruns of every experiment, thereby strengthening reproducibility.

\subsection{Data collection and preprocessing}
\label{sec:data}

We collected publicly available YouTube comments using the YouTube Data API v3, in accordance with the platform’s Terms of Service, with ethical approval from the University of xxx (institutional details withheld for double-blind review). Comments were retrieved from four science education channels—Museum of Science, SciShow, Sick Science!, and World Science Festival—spanning 2022–2024 and including both \textit{Videos} and \textit{Shorts}. Channels were selected for their broad content and large subscriber bases (>750{,}000) with extensive video libraries (>900 uploads), ensuring a diverse comment pool.

The dataset reflects moderately structured educational discourse characterised by relatively high content quality and sustained audience engagement, rather than the full spectrum of YouTube interactional styles. We collected 2{,}198 videos, yielding 609{,}763 top-level comments and 154{,}180 replies. All collection scripts, channel identifiers, and cut-off dates will be provided in an anonymised GitHub repository upon acceptance.

For external validation, we sampled four additional learning channels in Language (BBC Learning English), Medicine (Intellect Medicos), Music (Learn Piano with Jazer Lee), and Programming (Programming with Mosh), totalling 42{,}152 comments. These domains exhibit distinct epistemic and communicative profiles, providing a robust test of model generalisation.

Preprocessing involved removing duplicate comments, lowercasing, compressing whitespace, and retaining discourse-relevant punctuation. Tokenisation used SentencePiece to ensure compatibility with the DeBERTa backbone, and sequences were truncated or padded to 256 tokens. 

To prevent content leakage, cross-validation folds were assigned at the video level, so all comments from a given video were used exclusively for training or validation. Ten folds were stratified by video format (Shorts vs.\ regular Videos) to ensure balanced evaluation and reliable estimates of model performance.

\subsection{Knowledge construction coding}
\label{sec:kc-coding}

Prior to annotation, three coders were instructed using the codebook (see Table~\ref{tab:knowledge-construction1} and ~\ref{tab:knowledge-construction2}). They then independently annotated a random sample of 1,000 platform comments presented in random order, without any discussion among them. The resulting Cohen’s~$\kappa$ was 0.67. Coders were encouraged to provide brief written notes for borderline cases. Following a discussion of discrepancies, boundary conditions were incorporated into the codebook (see Table~\ref{tab:coding_book1} and Table~\ref{tab:coding_book2} in the appendix). The coders then independently annotated a second random sample of 1,000 comments, drawn from a different set than the first round, after which the Cohen’s~$\kappa$ increased to 0.88.

To construct the main annotated dataset, we first addressed the substantial class imbalance in the initial sample. Because \textit{NonKC} and \textit{Share} occurred far more frequently than \textit{Explore} and \textit{Integrate}, coder~A continued annotating until the minority categories reached 6{,}000 instances, resulting in a preliminarily oversampled dataset. Subsequently, comments with duplicate content were removed, and a subset of 5{,}700 comments per category was created. All labels assigned by coder~A were then removed, and coders~B and~C independently re-annotated these comments. At this stage, inter-rater reliability improved to a Cohen’s~$\kappa$ of 0.79, indicating a high level of agreement. When coders~B and~C reached agreement, their label was retained as final; when disagreement occurred, the label matching coder~A’s earlier annotation was adopted. Remaining disagreements were resolved through discussion. Finally, a balanced random sample of 5{,}000 comments per category (20{,}000 comments in total) was drawn to form the main analysis set.

After completing the main annotated dataset, we applied the same codebook to create test sets for external cross-domain validation. During this process, coders B and C were informed of the learning domain of each video to prevent domain-specific content features from being misclassified. Each domain contributed two subsets. First, 600 comments were randomly sampled and double-annotated to form a random test set reflecting the natural distribution of knowledge-construction categories. Second, we targeted at least 100 instances of each category in every domain by oversampling underrepresented categories from the remaining comments; when a category contained fewer than 100 comments, all available instances were included. Classifier predictions were used to identify likely candidates, which were then annotated manually. This procedure yielded an augmented subset with sufficient coverage for per-category analysis. 

While oversampling ensures adequate per-category coverage, it necessarily alters the natural distribution of knowledge-construction categories within each domain. As such, these augmented sets should be interpreted as diagnostic test beds rather than representative domain corpora. This design choice strengthens per-class evaluation but reduces comparability with real-world class frequencies, a trade-off common in educational NLP when minority epistemic categories are infrequent \parencite{Neto2021,Hu2021,Ba2023}.   

As noted by \textcite{Gunawardena2023}, a single comment may reflect more than one developmental stage. However, the difference between assigning multiple stages and assigning only the highest stage is not statistically significant. Based on this finding, \textcite{Gunawardena2023} argued that multilabel annotation is unnecessary when sufficient training data are available. Accordingly, in this study, each comment in the annotated dataset was assigned the label corresponding to the highest applicable category.

\subsection{Models and baselines}
\label{sec:baselines}

\begin{figure*}[htbp]
    \centering
    \includegraphics[width=0.72\linewidth]{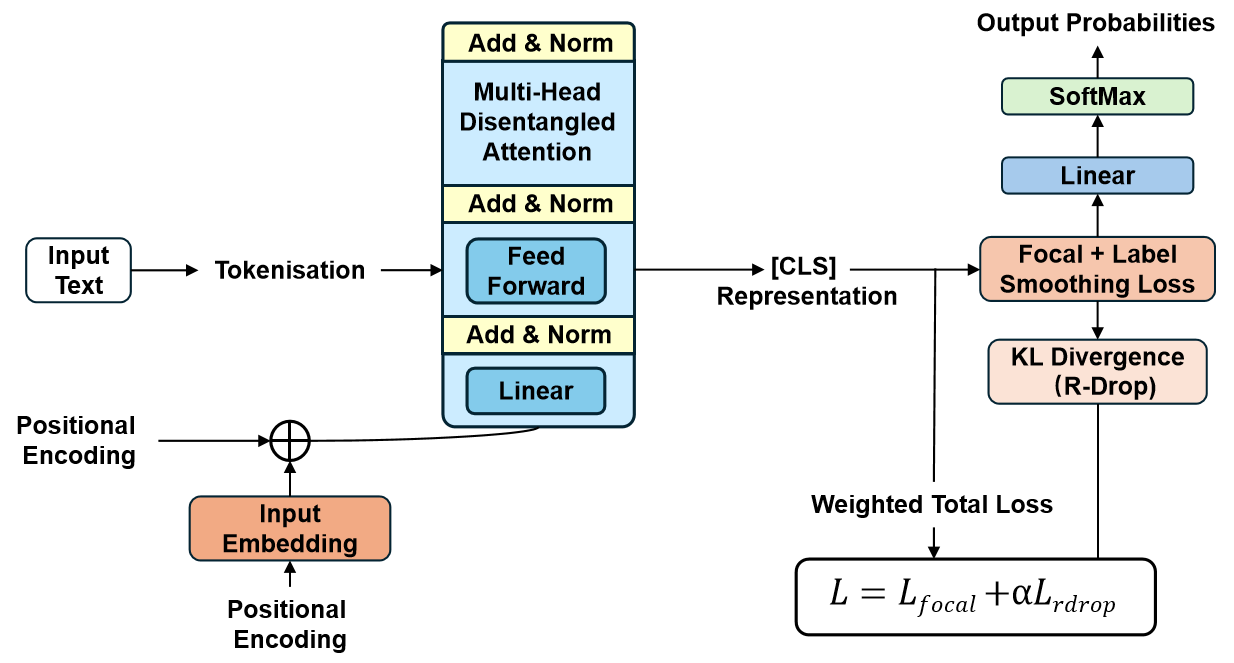}
    \caption{Proposed DeBERTa-KC model.}
    \label{fig:model_structure}
\end{figure*}

We benchmark \textbf{DeBERTa-KC} against two classical and three transformer baselines: TF--IDF (character $n$-grams 3--5) + Logistic Regression (balanced class weights) and TF--IDF + LinearSVC,  both implemented with scikit-learn \parencite{pedregosa2011scikit}; DistilBERT-base \parencite{sanh2019distilbert}, RoBERTa-base \parencite{liu2019roberta}, and DeBERTa-v3-base \parencite{he2021debertav3}. All transformer models are implemented using the Hugging Face Transformers library \parencite{wolf2020transformers}, and all models share identical tokenisation, fold assignments, and optimisation protocol for comparability. The proposed \textbf{DeBERTa-KC} architecture builds upon the \textit{DeBERTa-v3-large} backbone. Figure~\ref{fig:model_structure} illustrates the overall architecture.

\subsubsection{Backbone encoder and classification head}
\label{sec:backbone}

We employ \textbf{microsoft\addslash deberta-v3-large}, a transformer with 24 layers, 16 self-attention heads per layer, and hidden size 1024. DeBERTa disentangles content and positional embeddings and applies a relative position bias in self-attention, improving contextual representation over BERT and RoBERTa for fine-grained discourse signals \parencite{He2021}.

Each input comment is tokenised (SentencePiece), truncated or padded to a maximum of 256 tokens, and fed to a DeBERTa-v3-large encoder with the standard dropout configuration (dropout probability $p = 0.1$ in the encoder and classification head during training). We use the final-layer \texttt{[CLS]} representation $h_{\text{CLS}}$ as a sequence summary and, during training, apply this dropout before the linear classification head to reduce overfitting; dropout is disabled at inference time:
\[
\mathbf{z} = \text{Dropout}(h_{\text{CLS}}), \qquad 
\mathbf{p} = \text{Softmax}(\mathbf{W}\mathbf{z} + \mathbf{b}),
\]
where $\mathbf{W} \in \mathbb{R}^{4\times 1024}$ and $\mathbf{b} \in \mathbb{R}^{4}$. The vector $\mathbf{p}$ yields class probabilities for the four KC categories.

A compact head (single linear layer) minimises additional parameters, reducing overfitting on rare but deeper knowledge-construction labels  (e.g., \textit{Integrate})  while preserving transparency, as the head’s logits can be inspected or temperature-scaled for calibrated advisory use in learning analytics tools.

\subsubsection{Training objective}
\label{sec:objective}

To improve robustness under imbalance and promote trustworthy probabilities, the loss integrates three complementary components: \textit{Focal Loss}, \textit{Label Smoothing}, and \textit{R-Drop}. Together they address (a) class-frequency skew, (b) overconfident predictions, and (c) stochastic inconsistency.

We extend cross-entropy with Focal weighting \parencite{Lin2020} and Label Smoothing \parencite{Szegedy2016}. For gold label $y$ and predicted probabilities $\mathbf{p}$:
\[
\mathcal{L}_{\text{FL}} = - \left(1 - p_y\right)^{\gamma} \log\!\left(p'_y\right), \qquad
p'_y = (1-\epsilon)\,p_y + \frac{\epsilon}{K},
\]
with focusing parameter $\gamma = 2.0$, smoothing $\epsilon = 0.05$, and $K=4$ classes. Focal Loss down-weights easy examples and gives more gradient signal to hard cases (often \textit{Explore} or \textit{Integrate}), while smoothing curbs overconfidence for better-calibrated probabilities.

R-Drop encourages consistency between two stochastic forward passes of the same input by minimising bidirectional KL divergence \parencite{Liang2021}. For two predicted distributions $\mathbf{p}_1$ and $\mathbf{p}_2$:
\[
\mathcal{L}_{\text{RD}} = \tfrac{1}{2}\Big[\mathrm{KL}(\mathbf{p}_1\|\mathbf{p}_2) + \mathrm{KL}(\mathbf{p}_2\|\mathbf{p}_1)\Big].
\]
This reduces variance due to dropout and attention noise, which is important when model outputs underpin educator workflows (e.g., alerting on threads needing facilitation).

The final loss is an unweighted sum of the focal-loss and R-Drop terms:
\[
\mathcal{L} = \mathcal{L}_{\text{FL}} + \lambda_{\text{RD}}\,\mathcal{L}_{\text{RD}}, \qquad \lambda_{\text{RD}} = 1.0.
\]
We fix $\lambda_{\text{RD}}$ to 1.0, rather than tuning it, to keep the configuration simple, and backpropagate gradients through the encoder and classification head jointly.

\subsubsection{Training procedure}
\label{sec:traning}

We fine-tune with \texttt{AdamW} and cosine decay (initial learning rate $1\times 10^{-5}$, weight decay $0.05$, warm-up ratio $0.1$), early stopping (patience $=2$), and mixed precision (FP16/BF16). Batch sizes are 8 (train) and 16 (evaluation). Stratified 10-fold cross-validation ensures balanced label representation.

At inference, we form an ensemble by averaging the predicted class probabilities across folds, using simple (unweighted) averaging so that all fold models are treated symmetrically, and the combination method remains transparent and easily comparable across experiments:

\[
\hat{\mathbf{p}} = \frac{1}{N}\sum_{i=1}^{N}\mathbf{p}^{(i)}, \qquad N=10.
\]
The predicted label is $\arg\max_c \hat{p}_c$. Ensembling reduces variance and stabilises per-class metrics across folds. Alongside $\hat{\mathbf{p}}$, we compute predictive entropy $H(\hat{\mathbf{p}}) = -\sum_c \hat{p}_c \log \hat{p}_c$ as a lightweight uncertainty score.

We implement the model in PyTorch via HuggingFace Transformers, enabling reproducible configuration of tokenisation, truncation length (256), optimiser, and schedulers. All random seeds are fixed; training and evaluation artefacts (metrics per fold, confusion matrices, hyperparameters) are versioned.

\subsubsection{Evaluation protocol and metrics}
\label{sec:evaluation}

We use \emph{stratified, grouped-by-video} 10-fold cross-validation on the 20{,}000-item development corpus: in each run, we train on nine folds and evaluate on one, rotating across folds. We report mean~$\pm$~SD across folds for accuracy, macro-F1, weighted-F1, and per-class precision/recall/F1.

Hyperparameters are fixed across models; no tuning is performed on the external sets. At inference, we average fold-wise probabilities (simple ensemble) to stabilise predictions and optionally surface predictive entropy as an uncertainty signal for educator-facing use (not used to optimise metrics).

For summary metrics and per-class F1, we compute non-parametric bootstrap 95\% confidence intervals (2{,}000–5{,}000 resamples, i.i.d.). When a video ID is available, we additionally provide cluster (block) bootstrap CIs by resampling videos to account for intra-video correlation.

\subsection{Statistical analysis}
\label{sec:stats}

We treated the 10 cross-validation folds as repeated measurements for each model and computed macro-F1 scores per fold. This within-subjects design evaluates all models on the same data splits. Because fold-level scores ($N=10$) cannot be assumed to be normally distributed, we used non-parametric tests. A Friedman test assessed overall differences among models, followed by Wilcoxon signed-rank tests for pairwise comparisons. $p$-values were adjusted using the Holm method ($\alpha=.05$). For each model, we report the mean macro-F1, standard deviation, and non-parametric bootstrap 95\% confidence intervals. For pairwise comparisons, we additionally report mean differences ($\Delta$) and paired-sample Cohen’s $d$. Linear mixed-effects models were considered, but fold-level performance scores contain only ten observations per model, which limits the stability of variance-component estimation and leads to unreliable random-effects structures. Given this constraint, non-parametric repeated-measures tests offer a more appropriate and conservative alternative for evaluating model differences.

\section{Results}
\label{sec:results}

\subsection{RQ1: Reliability of the Adapted IAM Coding Scheme}
\label{sec:RQ1}

\begin{table*}[htbp]
\centering
\caption{Knowledge construction categories (Part 1).}
\label{tab:knowledge-construction1}
\begin{tabularx}{\textwidth}{p{6cm}| p{4cm}| X}
\toprule
\textbf{Gunawardena et al. (1997)} & \textbf{Nguyen \& Diederich (2023)} & \textbf{Category Indicator} \\
\midrule
N/A & \textbf{Non-knowledge construction (NonKC)} \newline
Comment to socialise (positive and negative reactions), with less focus on the video’s content. This code is added to capture sentiments. & \textbf{NonKC}
\begin{itemize}
  \item Express emotion
  \item Socialise or acknowledge others
  \item Produce irrelevant or incoherent content
  \item Ask unrelated questions to the learning content
\end{itemize} \\

\midrule
\textbf{PhI: Sharing/comparing of information} \newline
PhI/A. A statement of observation or opinion \newline
PhI/B. A statement of agreement from one or more other participants \newline
PhI/C. Corroborating examples provided by one or more participants \newline
PhI/D. Asking and answering questions to clarify details of statements \newline
PhI/E. Definition, description, or identification of a problem
& \textbf{Share ideas}\newline
Ask clarifying questions, seek information or provide simple statements (personal experiences, facts or opinions). Comments are related to video content.
&
\textbf{Share}
\begin{itemize}
    \item State an opinion or observation (PhI/A)
    \item Express simple agreement (PhI/B)
    \item Provide simple example or information (PhI/C)
    \item Ask for clarification (PhI/D)
    \item Identify a problem or issue (PhI/E)
\end{itemize}
\\

\midrule
\textbf{PhII: The discovery and exploration of dissonance or inconsistency among ideas, concepts or statements}\newline
PhII/A. Identifying and stating areas of disagreement\newline
PhII/B. Asking and answering questions to clarify the source and extent of disagreement\newline
PhII/C. Restating the participant's position, and possibly advancing arguments or considerations in its support by references to the participant's experience, literature, formal data collected, or proposal of relevant metaphor or analogy to illustrate point of view
& 

\textbf{Explore dissonances} \newline
State agreement or disagreement (including simple statements such as ‘I agree’). Ask questions to clarify the extent of disagreement.

&
\textbf{Explore}
\begin{itemize}
    \item Express disagreement (PhII/A)
    \item Question or probe another statement (PhII/B)
    \item Defend one’s position with minimal support (PhII/C)
\end{itemize}
\\
\bottomrule
\end{tabularx}
\end{table*}

\begin{table*}[htbp]
\centering
\caption{Knowledge construction categories (Part 2).}
\label{tab:knowledge-construction2}
\begin{tabularx}{\textwidth}{p{6cm}| p{4cm}| X}
\toprule
\textbf{Gunawardena et al. (1997)} & \textbf{Nguyen \& Diederich (2023)} & \textbf{Category Indicator} \\
\midrule
\textbf{PhIII: Negotiation of meaning/co-construction of knowledge} \newline
PhIII/A. Negotiation or clarification of the meaning of terms \newline
PhIII/B. Negotiation of the relative weight to be assigned to types of argument \newline
PhIII/C. Identification of areas of agreement or overlap among conflicting concepts \newline
PhIII/D. Proposal and negotiation of new statements embodying compromise, co-construction \newline
PhIII/E. Proposal of integrating or accommodating metaphors or analogies

& \textbf{Negotiate} \newline
Clarify concepts. Propose and negotiate areas of disagreement to integrate ideas. Employ more extensive evidence and explanation than earlier phases.

&
\textbf{Integrate}
\begin{itemize}
    \item Clarify concepts or terms (PhIII/A)
    \item Propose or synthesize integrative statements (PhIII/C, PhIII/D, PhIII/E, PhV/A)
    \item Evaluate statements by comparing perspectives, experiences, or information sources (PhIII/B, PhIV/A, PhIV/B, PhIV/C, PhIV/D, PhIV/E)
    \item Apply or reflect on new understanding (PhV/B, PhV/C)
\end{itemize}
\\

\cline{1-2}
\textbf{PhIV: Testing and modification of proposed synthesis or co-construction} \newline
PhIV/A. Testing the proposed synthesis against "received fact" as shared by the participants and/or their culture  \newline
PhIV/B. Testing against existing cognitive schema  \newline
PhIV/C. Testing against personal experience  \newline
PhIV/D. Testing against formal data collected  \newline
PhIV/E. Testing against contradictory testimony in the literature
& \textbf{Test/modify proposed ideas}\newline
Test the proposed idea syntheses against other contexts (data, references, personal experiences).

& 
\\
\cline{1-2}
\textbf{PhV: Agreement statement(s)/applications of newly-constructed meaning}\newline
PhV/A. Summarization of agreement(s)\newline
PhV/B. Applications of new knowledge\newline
PhV/C. Metacognitive statements by the participants illustrating their understanding that their knowledge or ways of thinking (cognitive schema) have changed as a result of the conference interaction
& \textbf{Apply new knowledge}\newline
Summarise agreement, make reflective statements and apply new knowledge. 
&  \\
\bottomrule
\end{tabularx}
\end{table*}

We draw on the original IAM \parencite{Gunawardena1997} and its adaptation for social media contexts by \textcite{Nguyen2023} to develop a codebook suited to non-linear discourse environments (Table~\ref{tab:knowledge-construction1} and ~\ref{tab:knowledge-construction2}). Following \textcite{Nguyen2023}, we added the category non-knowledge construction (NonKC) to capture comments unrelated to learning processes. This addition is necessary because social media platforms contain large volumes of comments that are social, humorous, emotional, phatic, or otherwise not oriented toward epistemic engagement, and failing to account for these responses would bias the distribution of knowledge-construction categories and inflate model estimates. For knowledge-construction discourse, we retained all indicators from IAM Phases~I and~II and reorganised them into two categories: \textit{Share} and \textit{Explore}. Previous research has shown that higher-level IAM Phases~IV and~V rarely appear in online comments; removing or merging these phases into Phase~III has been shown to produce more robust model performance \parencite{Gunawardena2023,Nguyen2023}. Accordingly, we combined these three phases into a single category, \textit{Integrate}, and developed corresponding indicators based on the subcomponents specified in IAM Phases~III--V. The final framework thus consists of four categories: NonKC, Share, Explore, and Integrate. This adaptation transforms IAM from a framework describing linear, iterative developmental phases into one that characterises discrete, non-sequential discourse categories at the comment level.

In the main annotated dataset of 20,000 comments, the application of this codebook achieved an inter-rater reliability of Cohen's $\kappa = 0.79$, indicating an almost perfect level of agreement \parencite{landis1977measurement}. We then examined coder agreement in the complete random subsets from four external domains. Across these domains, Cohen's $\kappa$ values ranged from .853 to .926, likewise falling within the ``almost perfect'' range. In addition to $\kappa$, annotation accuracy ranged from .913 to .958 and macro-F1 scores from .876 to .949 (see Table~\ref{tab:rq3_overall_normal}). These reliability estimates are based on a balanced sample of independently coded comments and do not adjust for clustering by video or domain; as such, they quantify agreement at the comment level rather than providing a cluster-adjusted $\kappa$ for the natural, imbalanced distributions. This demonstrates that the codebook provides stable and reliable annotations across domains with markedly different discourse characteristics, indicating that IAM can indeed be adapted for knowledge construction assessment in noninteractive comments. These findings lay the foundation for the supervised training and evaluation of the automatic classifier reported in RQ2, as well as the cross-domain generalisation analyses presented in RQ3.

\subsection{RQ2: Comparison of Classifiers for IAM-Based Knowledge Construction}
\label{sec:RQ2}

\subsubsection{Model performance and comparisons}\label{sec:overall}

\begin{table}[htbp]
\centering
\caption{Overall performance (10-fold CV). Values are mean $\pm$ SD. The best model (based on mean Accuracy) is shown in bold.}
\resizebox{\linewidth}{!}{
\begin{tabular}{lccc}
\hline
\textbf{Model} & \textbf{Accuracy} & \textbf{Macro-F1} & \textbf{Weighted-F1} \\
\hline
TF-IDF + Logistic Regression & .729 $\pm$ .012 & .728 $\pm$ .012 & .728 $\pm$ .012 \\
TF-IDF + LinearSVC & .711 $\pm$ .009 & .709 $\pm$ .009 & .709 $\pm$ .009 \\
DistilBERT (base) & .807 $\pm$ .010 & .807 $\pm$ .010 & .807 $\pm$ .010 \\
RoBERTa-base & .821 $\pm$ .010 & .821 $\pm$ .009 & .821 $\pm$ .009 \\
DeBERTa-v3-base & .834 $\pm$ .007 & .834 $\pm$ .007 & .834 $\pm$ .007 \\
DeBERTa-v3-large (base backbone) & .833 $\pm$ .010 & .834 $\pm$ .010 & .834 $\pm$ .010 \\
\textbf{DeBERTa-v3-large \textit{(ours: +Focal +LS +R-Drop)}} & \textbf{.841} $\pm$ .011 & \textbf{.841} $\pm$ .010 & \textbf{.841} $\pm$ .010 \\
\quad – no R-Drop & .833 $\pm$ .011 & .834 $\pm$ .010 & .834 $\pm$ .010 \\
\quad – no LS & .840 $\pm$ .009 & .841 $\pm$ .009 & .841 $\pm$ .009 \\
\quad – no Focal & .836 $\pm$ .008 & .836 $\pm$ .008 & .836 $\pm$ .008 \\
\hline
\end{tabular}}
\label{tab:overall_cv}
\end{table}

Table~\ref{tab:overall_cv} summarises the overall performance of all models in terms of mean accuracy, macro-F1 and weighted-F1 over 10-fold cross-validation. The two TF--IDF baselines show substantially lower performance (macro-F1 $\approx 0.71$--$0.73$) than all transformer-based models (macro-F1 $\approx 0.81$--$0.84$). This pattern indicates that, for our dataset of student contributions, pre-trained language models capture relevant linguistic and contextual cues more effectively than traditional bag-of-words approaches.

Among the transformer-based models, the DeBERTa variants achieve the best results and are tightly clustered in a narrow performance band. Our best configuration, \emph{DeBERTa-v3-large (ours: +Focal +LS +R-Drop)}, attains the highest mean macro-F1 ($0.841 \pm 0.010$), with a bootstrap 95\% confidence interval of approximately $[0.835, 0.848]$. The ablation variants, where we removed R-Drop, label smoothing, or focal loss, show only small decreases in macro-F1 (on the order of $0.005$--$0.008$), and \emph{DeBERTa-v3-base} is also close to the best configuration. Although we highlight the full DeBERTa-v3-large configuration as the best model in terms of mean accuracy and macro-F1, the performance differences among DeBERTa variants are modest.

\begin{figure}[htbp]
    \centering
    \includegraphics[width=\linewidth]{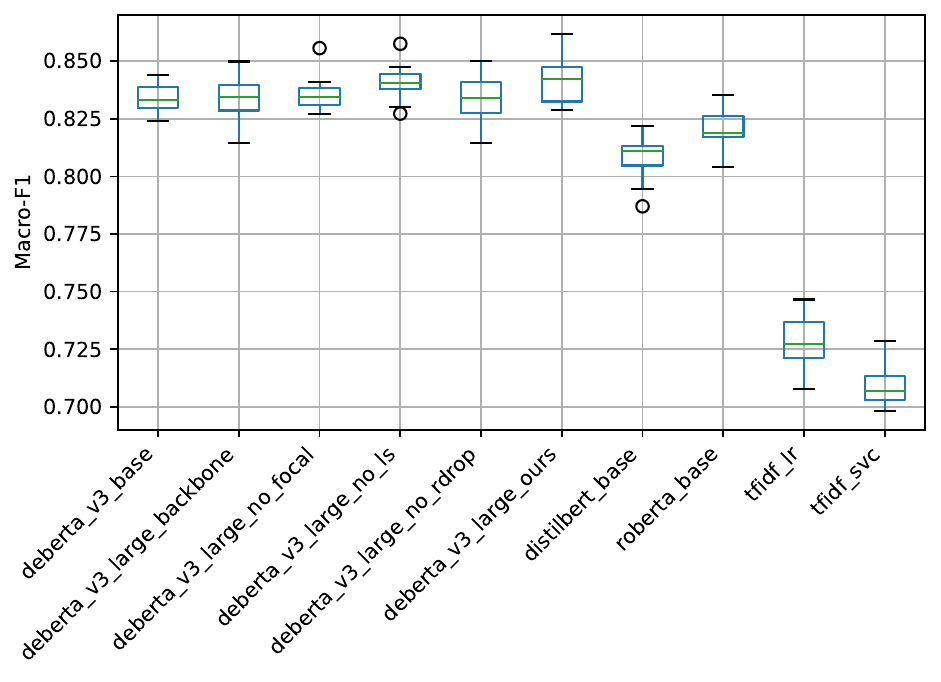}
    \caption{Distribution of 10-fold macro-F1 scores for all models. Each box shows the variability across folds, with individual fold scores overlaid as points. Transformer-based models consistently outperform the TF--IDF baselines, and the DeBERTa variants form a compact group of high-performing models.}
    \label{fig:macro_f1_boxplot}
\end{figure}

\begin{figure}[htbp]
    \centering
    \includegraphics[width=\linewidth]{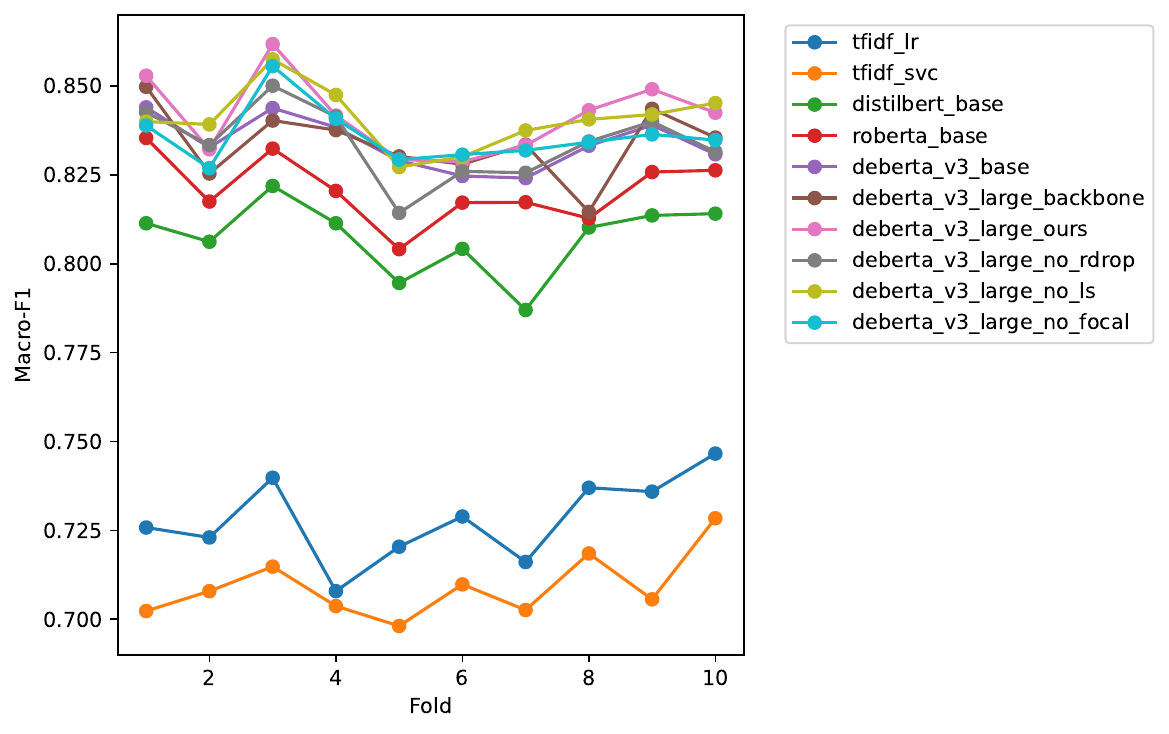}
    \caption{Fold-wise macro-F1 trajectories for all models. Each line corresponds to a model evaluated on the same 10 cross-validation folds. Transformer-based models dominate the TF--IDF baselines on nearly every fold, illustrating the repeated-measures structure of the analysis.}
    \label{fig:macro_f1_lines}
\end{figure}

Figure~\ref{fig:macro_f1_boxplot} visualises the distribution of 10-fold macro-F1 scores for each model. The TF--IDF baselines exhibit consistently lower scores and little overlap with the transformer models. In contrast, the transformer models form a compact group of high-performing models, with the DeBERTa variants showing both high central performance and relatively low variability across folds. Figure~\ref{fig:macro_f1_lines} further illustrates that, on almost every fold, transformer models outperform the TF--IDF baselines when evaluated on the same data splits, reinforcing the repeated-measures design of our analysis.

The Friedman test on fold-level macro-F1 scores across all models revealed a highly significant overall effect, $\chi^2(9) = 77.38,\ p < 10^{-12}$, indicating systematic differences in performance. We therefore conducted pairwise Wilcoxon signed-rank tests with Holm correction. All transformer-based models significantly outperformed both TF--IDF baselines after correction for multiple comparisons. For example, the contrast between \emph{DeBERTa-v3-large (ours)} and TF--IDF + Logistic Regression yielded an average macro-F1 difference of $\Delta \approx 0.113$ in favour of our model, with a Holm-adjusted $p < .05$ and a large paired effect size. From an educational perspective, this means that transformer models are much more reliable in assigning the correct knowledge construction category to student contributions than the simpler TF--IDF approaches.

Differences among the transformer models themselves were smaller. Although \emph{DeBERTa-v3-large (ours)} achieved slightly higher mean macro-F1 than \emph{DeBERTa-v3-base} and \emph{RoBERTa-base} (e.g., $\Delta \approx 0.008$ and $\Delta \approx 0.021$, respectively), these improvements did not remain statistically significant after Holm correction. Likewise, the ablated DeBERTa-v3-large variants (without R-Drop, label smoothing, or focal loss) did not differ significantly from the full model. In practical terms, our full DeBERTa configuration is numerically the best-performing model, but its advantage over other DeBERTa variants is modest; the most robust result is the large and statistically significant performance gap between any transformer-based model and the TF--IDF baselines. Consequently, we adopt the full DeBERTa-v3-large configuration as our primary model for subsequent analyses, not because its performance differences are statistically decisive, but because it consistently attains the highest macro-F1 and exhibits the most stable and well-calibrated probability estimates across folds.

\subsubsection{Per-category performance}
\label{sec:per-class}

To better understand how the classifiers behave with respect to the different categories of knowledge construction, we analysed precision, recall, and F1 for each category separately (Tables~\ref{tab:classD}--\ref{tab:classA}). Across all four categories, transformer-based models consistently outperform the TF--IDF baselines, but the relative difficulty of the categories and the benefits of the DeBERTa variants differ slightly by category.

For the NonKC category  (Table~\ref{tab:classD}), all models perform relatively well, which is expected given that nonKC comments often contain clearer superficial or off-topic cues.  The TF--IDF baselines still lag behind (F1 $\approx .72$--.74), but the gap to transformers is somewhat smaller than for the knowledge-building categories. Within the transformer family, DeBERTa-v3-base has the highest precision (.883 $\pm$ .024), whereas the ``no LS'' variant attains the highest F1 (.846 $\pm$ .009). Our full model strikes a balance, with slightly lower precision (.861 $\pm$ .032) but the highest recall across all models (.827 $\pm$ .038), yielding an F1 of .843 $\pm$ .017. This pattern suggests that the full configuration conservatively avoids misclassifying knowledge-construction comments as NonKC, which is advantageous when the model is used to identify meaningful knowledge-construction episodes.

\begin{table}[htbp]
\centering
\caption{Performance comparison across models for the \textbf{NonKC} category.}
\resizebox{\linewidth}{!}{
\begin{tabular}{lccc}
\hline
\textbf{Model} & \textbf{Precision} & \textbf{Recall} & \textbf{F1} \\
\hline
TF-IDF + Logistic Regression & .734 $\pm$ .020 & .736 $\pm$ .026 & .735 $\pm$ .019 \\
TF-IDF + LinearSVC & .714 $\pm$ .021 & .725 $\pm$ .023 & .719 $\pm$ .018 \\
DistilBERT (base) & .831 $\pm$ .015 & .782 $\pm$ .029 & .805 $\pm$ .013 \\
RoBERTa-base & .858 $\pm$ .024 & .782 $\pm$ .044 & .817 $\pm$ .018 \\
DeBERTa-v3-base & \textbf{.883} $\pm$ .024 & .782 $\pm$ .029 & .829 $\pm$ .011 \\
DeBERTa-v3-large (replaced with base backbone) & .878 $\pm$ .022 & .790 $\pm$ .025 & .831 $\pm$ .012 \\
DeBERTa-v3-large \textit{(ours: +Focal +LS +R-Drop)} & .861 $\pm$ .032 & \textbf{.827} $\pm$ .038 & .843 $\pm$ .017 \\
\quad \;– no R-Drop & .873 $\pm$ .022 & .802 $\pm$ .031 & .836 $\pm$ .013 \\
\quad \;– no LS & .869 $\pm$ .033 & .826 $\pm$ .035 & \textbf{.846} $\pm$ .009 \\
\quad \;– no Focal & .873 $\pm$ .033 & .808 $\pm$ .036 & .838 $\pm$ .012 \\
\hline
\end{tabular}}
\label{tab:classD}
\end{table}

The Share category (Table~\ref{tab:classB}) is more challenging for all models and shows the lowest scores among the knowledge construction categories. The TF--IDF baselines perform markedly worse here (F1 $\approx .62$--.65), whereas transformer models yield F1 scores in the mid-.70s to high-.79s. Our full DeBERTa configuration achieves an F1 of .799 $\pm$ .014 with balanced precision (.796 $\pm$ .029) and recall (.804 $\pm$ .030). The ``no LS'' variant obtains a very similar F1 (.797 $\pm$ .016) with slightly higher precision but slightly lower recall. DeBERTa-v3-base yields the highest recall for Share (.817 $\pm$ .024), suggesting that label smoothing may help the model recover more borderline sharing contributions at the cost of a small precision loss. These results indicate that distinguishing Share from the other categories is inherently challenging at both the conceptual and linguistic levels, which may partly reflect the fact that the operational definition of Share in this study encompasses a broader and more diverse set of behaviors than the other categories.

\begin{table}[htbp]
\centering
\caption{Performance comparison across models for the \textbf{Share} category.}
\resizebox{\linewidth}{!}{
\begin{tabular}{lccc}
\hline
\textbf{Model} & \textbf{Precision} & \textbf{Recall} & \textbf{F1} \\
\hline
TF-IDF + Logistic Regression & .652 $\pm$ .021 & .646 $\pm$ .023 & .648 $\pm$ .017 \\
TF-IDF + LinearSVC & .634 $\pm$ .016 & .614 $\pm$ .027 & .623 $\pm$ .018 \\
DistilBERT (base) & .752 $\pm$ .022 & .782 $\pm$ .032 & .766 $\pm$ .017 \\
RoBERTa-base & .768 $\pm$ .025 & .785 $\pm$ .027 & .775 $\pm$ .011 \\
DeBERTa-v3-base & .772 $\pm$ .015 & \textbf{.817} $\pm$ .024 & .794 $\pm$ .009 \\
DeBERTa-v3-large (replaced with base backbone) & .781 $\pm$ .022 & .800 $\pm$ .021 & .790 $\pm$ .013 \\
DeBERTa-v3-large \textit{(ours: +Focal +LS +R-Drop)} & \textbf{.796} $\pm$ .029 & .804 $\pm$ .030 & \textbf{.799} $\pm$ .014 \\
\quad \;– no R-Drop & .777 $\pm$ .027 & .804 $\pm$ .037 & .789 $\pm$ .013 \\
\quad \;– no LS & .793 $\pm$ .032 & .803 $\pm$ .049 & .797 $\pm$ .016 \\
\quad \;– no Focal & .775 $\pm$ .020 & .812 $\pm$ .020 & .793 $\pm$ .009 \\
\hline
\end{tabular}}
\label{tab:classB}
\end{table}

For the Explore category (Table~\ref{tab:classC}), performance again improves substantially when moving from TF--IDF (F1 $\approx .66$--.69) to transformer-based models (F1 $\approx .77$--.82). Among transformers, the DeBERTa variants again dominate. Our full model achieves the highest F1 (.821 $\pm$ .016), with the best precision (.803 $\pm$ .032) and strong recall (.842 $\pm$ .026). The ``no LS'' variant achieves the highest recall (.850 $\pm$ .030) and an F1 of .820 $\pm$ .015, essentially matching the full model. These patterns suggest that the additional regularisation from focal loss, label smoothing, and R-Drop helps DeBERTa-v3-large to capture exploratory moves slightly more reliably, without sacrificing precision. 


\begin{table}[htbp]
\centering
\caption{Performance comparison across models for the \textbf{Explore} category.}
\resizebox{\linewidth}{!}{
\begin{tabular}{lccc}
\hline
\textbf{Model} & \textbf{Precision} & \textbf{Recall} & \textbf{F1} \\
\hline
TF-IDF + Logistic Regression & .693 $\pm$ .022 & .679 $\pm$ .029 & .686 $\pm$ .023 \\
TF-IDF + LinearSVC & .679 $\pm$ .021 & .650 $\pm$ .022 & .664 $\pm$ .018 \\
DistilBERT (base) & .771 $\pm$ .034 & .776 $\pm$ .041 & .772 $\pm$ .016 \\
RoBERTa-base & .778 $\pm$ .021 & .816 $\pm$ .028 & .796 $\pm$ .016 \\
DeBERTa-v3-base & .789 $\pm$ .024 & .836 $\pm$ .029 & .812 $\pm$ .016 \\
DeBERTa-v3-large (replaced with base backbone) & .781 $\pm$ .033 & .845 $\pm$ .028 & .811 $\pm$ .016 \\
DeBERTa-v3-large \textit{(ours: +Focal +LS +R-Drop)} & \textbf{.803} $\pm$ .032 & .842 $\pm$ .026 & \textbf{.821} $\pm$ .016 \\
\quad \;– no R-Drop & .785 $\pm$ .034 & .841 $\pm$ .023 & .812 $\pm$ .014 \\
\quad \;– no LS & .794 $\pm$ .025 & \textbf{.850} $\pm$ .030 & .820 $\pm$ .015 \\
\quad \;– no Focal & .802 $\pm$ .026 & .834 $\pm$ .036 & .817 $\pm$ .015 \\
\hline
\end{tabular}}
\label{tab:classC}
\end{table}

For the Integrate category (Table~\ref{tab:classA}), all transformer models achieve very high performance, with F1 scores around or above .89. The TF--IDF baselines lag behind (F1 $\approx .83$--.84), indicating that capturing negotiation moves requires richer contextual representations than simple bag-of-words features can provide. Within the transformer family, the DeBERTa variants obtain the strongest results. Our full configuration (\emph{DeBERTa-v3-large, +Focal +LS +R-Drop}) attains an F1 of .903 $\pm$ .012, matching the best-performing large-backbone variant and slightly exceeding DeBERTa-v3-base (F1 = .902 $\pm$ .013). The ``no LS'' ablation achieves the highest precision (.920 $\pm$ .023) for Integrate but trades this for somewhat lower recall (.883 $\pm$ .026), suggesting a more conservative classification that misses some genuine contributions. 

\begin{table}[htbp]
\centering
\caption{Performance comparison across models for the \textbf{Integrate} category.}
\resizebox{\linewidth}{!}{
\begin{tabular}{lccc}
\hline
\textbf{Model} & \textbf{Precision} & \textbf{Recall} & \textbf{F1} \\
\hline
TF-IDF + Logistic Regression & .833 $\pm$ .015 & .855 $\pm$ .013 & .844 $\pm$ .011 \\
TF-IDF + LinearSVC & .808 $\pm$ .011 & .854 $\pm$ .015 & .830 $\pm$ .009 \\
DistilBERT (base) & .885 $\pm$ .028 & .890 $\pm$ .028 & .887 $\pm$ .013 \\
RoBERTa-base & .890 $\pm$ .027 & \textbf{.902} $\pm$ .036 & .895 $\pm$ .014 \\
DeBERTa-v3-base & .906 $\pm$ .031 & .899 $\pm$ .030 & .902 $\pm$ .013 \\
DeBERTa-v3-large (replaced with base backbone) & .909 $\pm$ .025 & .899 $\pm$ .034 & \textbf{.903} $\pm$ .013 \\
DeBERTa-v3-large \textit{(ours: +Focal +LS +R-Drop)} & .917 $\pm$ .014 & .891 $\pm$ .031 & \textbf{.903} $\pm$ .012 \\
\quad \;– no R-Drop & .916 $\pm$ .024 & .885 $\pm$ .035 & .899 $\pm$ .013 \\
\quad \;– no LS & \textbf{.920} $\pm$ .023 & .883 $\pm$ .026 & .900 $\pm$ .006 \\
\quad \;– no Focal & .907 $\pm$ .033 & .888 $\pm$ .040 & .897 $\pm$ .012 \\
\hline
\end{tabular}}
\label{tab:classA} 
\end{table}

In summary, transformer-based models deliver consistent improvements across all knowledge-construction categories and clearly outperform TF–IDF baselines, with the DeBERTa-v3-large variants performing best. Share and Explore remain the most difficult to classify due to their conceptual overlap and linguistic variability, whereas NonKC and Integrate are comparatively clearer. Although performance differences among the DeBERTa configurations are small and not statistically significant, the full model shows the highest and most stable results. These findings indicate that modern transformer architectures can provide sufficiently accurate comment-level category labels for large-scale analyses of students’ knowledge-construction behaviors.

\subsection{RQ3: Cross-Domain Generalisation of DeBERTa-KC}
\label{sec:RQ3}

\subsubsection{Overall cross-domain performance on realistic samples}
\label{sec:RQ3_overall}

\begin{figure}[htbp]
    \centering
    \includegraphics[width=\linewidth]{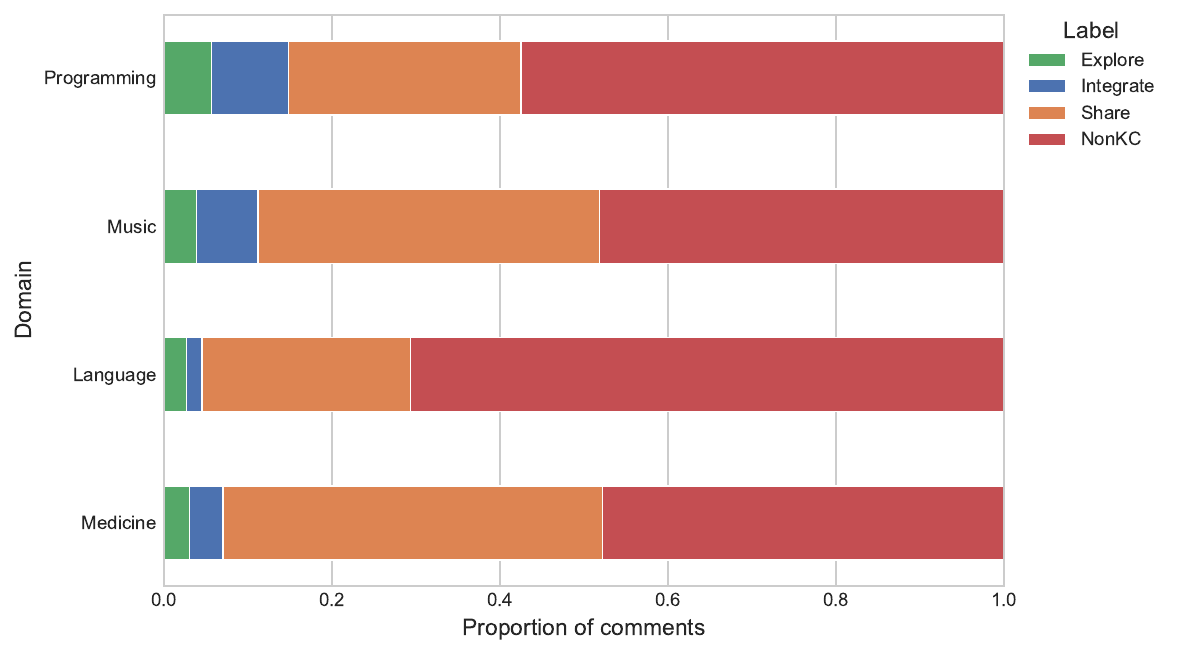}
    \caption{Natural knowledge construction category distribution by domain in the random subsets, based on coder~B's labels. Each horizontal bar shows the proportion of Explore, Integrate, Share, and NonKC comments within a domain.}
    \label{fig:rq3_label_distribution_normal}
\end{figure}

Figure~\ref{fig:rq3_label_distribution_normal} shows distribution of knowledge-construction categories in the Normal subset across four domains. In all domains, \textit{NonKC} is the most frequent category, followed by \textit{Share}, whereas \textit{Explore} and \textit{Integrate} occur far less often. In the language domain, \textit{NonKC} is dominant (71\%), \textit{Share} accounts for a moderate proportion (24.8\%), and both \textit{Explore} (2.7\%) and \textit{Integrate} (1.8\%) appear only rarely. In the medicine domain, \textit{NonKC} (47.8\%) remains slightly more prevalent than \textit{Share} (45.2\%), while \textit{Explore} (3.0\%) and \textit{Integrate} (4.0\%) show limited but noticeable presence. In the music domain, \textit{NonKC} has the largest \textit{share} (48.2\%), followed by \textit{Share} (40.7\%); \textit{Explore} (3.8\%) and \textit{Integrate} (7.3\%) occur somewhat more frequently than in the medicine domain. Finally, in the programming domain, \textit{NonKC} continues to be the largest category (57.5\%), \textit{Share} remains substantial (27.7\%), and \textit{Explore} (5.7\%) and \textit{Integrate} (9.2\%) reach their highest proportions across all domains. This pronounced skew, particularly in the language domain, also helps explain the lower macro-F1 scores observed there: domains with very few \textit{Explore} and \textit{Integrate} instances provide limited training and evaluation data for these stages, which depresses their per-category F1 and the overall macro-F1.

Table~\ref{tab:rq3_overall_normal} summarises overall classifier performance in each domain on the Normal subsets (600 comments per domain), using coder~B as the reference. Human agreement (coder~B vs.\ coder~C) on these same subsets is also shown as an approximate upper bound.

\begin{table*}[htbp]
\centering
\caption{Overall performance of \emph{DeBERTa-v3-large (ours)} and human agreement on the random subsets. Values are computed against coder~B (for the model) and between coder~B and coder~C (for human agreement).}
\resizebox{\linewidth}{!}{
\begin{tabular}{lrrrrrrr}
\hline
\textbf{Domain} & \textbf{$N$} & \textbf{Acc} (model vs B) & \textbf{Macro-F1} (model vs B) & \textbf{Weighted-F1} (model vs B) & \textbf{Acc} (B vs C) & \textbf{Macro-F1} (B vs C) & \textbf{$\kappa$} (B,C) \\
\hline
Medicine & 600 & .897 & .864 & .896 & .958 & .949 & .926 \\
Language & 600 & .820 & .707 & .803 & .940 & .893 & .865 \\
Music   & 600 & .752 & .705 & .738 & .948 & .923 & .913 \\
Programming & 600 & .882 & .818 & .881 & .913 & .876 & .853 \\
\hline
\end{tabular}}
\label{tab:rq3_overall_normal}
\end{table*}

Compared to the in-domain macro-F1 of approximately .841 reported in Section~\ref{sec:RQ2}, the classifier's macro-F1 on the Normal subsets is slightly higher in \textit{Medicine} (.864) and somewhat lower in \textit{Programming} domain (.818), yielding small absolute differences of $+0.023$ and $-0.023$, respectively. In contrast, performance drops more substantially in \textit{Language} and \textit{Music} domains, with macro-F1 scores of .707 and .705, corresponding to reductions of approximately $-0.13$ relative to the in-domain result. Part of this drop can be attributed to the natural distribution of categories in those domains, such as Language, which contain very few examples of \textit{Explore} and \textit{Integrate}, disproportionately affecting the macro-F1. In domains where higher-order epistemic moves are almost nonexistent, even small absolute errors translate into large proportional reductions in macro-level metrics. The weighted-F1 scores follow a similar pattern, reflecting the strong influence of the abundant \textit{NonKC} and \textit{Share} categories.

\begin{figure}[htbp]
    \centering
    \includegraphics[width=\linewidth]{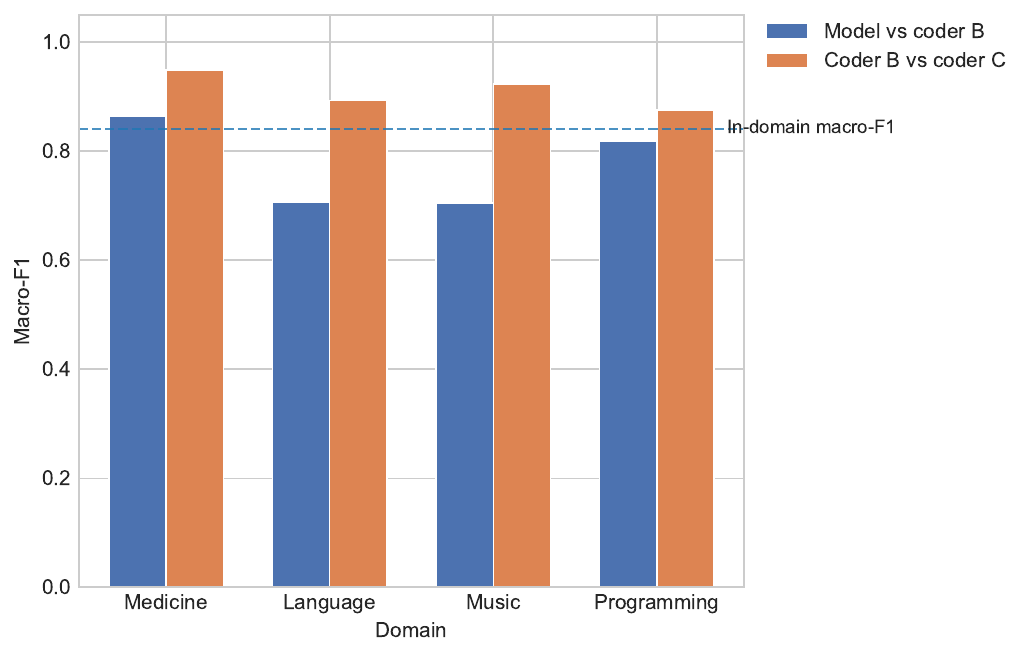}
    \caption{Cross-domain macro-F1 on the random subsets. Bars show macro-F1 for the classifier against coder~B and for coder~B against coder~C in each domain. The dashed line indicates the in-domain macro-F1 of the classifier on the original dataset (Section~\ref{sec:RQ2}).}
    \label{fig:rq3_macroF1_overall}
\end{figure}

Figure~\ref{fig:rq3_macroF1_overall} visualises these macro-F1 scores alongside the human--human macro-F1, with a dashed line indicating the in-domain macro-F1. The figure highlights two key points. First, in all domains there remains a sizeable gap between model--human agreement and human--human agreement (typically around .08--.14 macro-F1), indicating that the classifier has not yet reached human reliability. Second, cross-domain degradation is not uniform: the model retains most of its in-domain performance in Medicine and Programming domains, but exhibits noticeably weaker performance in Language and Music domains.

To examine whether classifier performance varied across domains, we fitted a logistic regression model in which the binary outcome indicated whether the model’s prediction matched Coder B’s annotation (correct vs.\ incorrect), with domain as a categorical predictor. A likelihood ratio test showed a significant domain effect ($\chi^2(3) = 56.81$, $p < .001$), indicating that the probability of correct prediction depended on the educational domain. Using Language domain as the reference category, the classifier was more likely to be correct in \textit{Medicine} ($\approx 1.90$) and \textit{Programming} ($\approx 1.64$) domains, but less likely in \textit{Music} ($\approx 0.66$) domain. 

\subsubsection{Per-category cross-domain generalisation}
\label{sec:RQ3_per_category}

\begin{figure}[htbp]
    \centering
    \includegraphics[width=\linewidth]{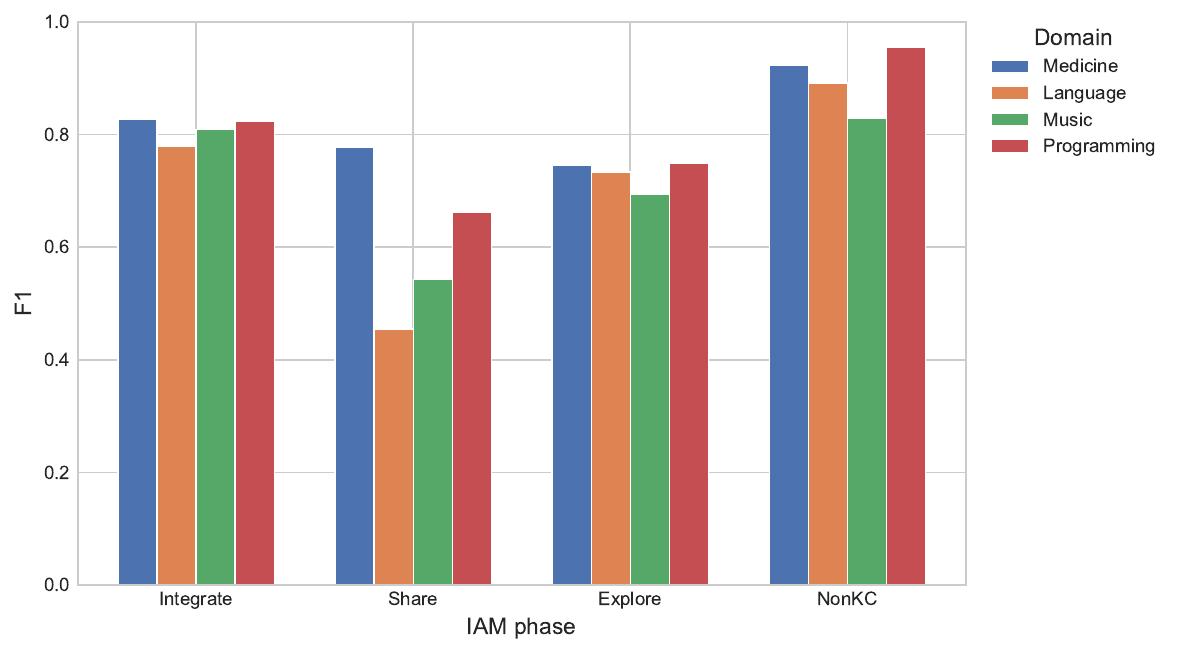}
    \caption{Per-category F1 scores by domain (all samples: Normal + Augment). Each group of bars corresponds to a knowledge construction category, with separate bars for Medicine, Language, Music, and Programming.}
    \label{fig:rq3_percategory_f1}
\end{figure}

To further examine which categories generalize well across domains and which remain challenging, we conducted a category-level analysis using the Normal and Augment subsets. Because the Augment subset oversampled previously rare stages, the resulting evaluation set does not reflect the natural class distribution but provides a more balanced basis for assessing the classifier's ability to identify each category across domains. Using all available comments (Normal + Augment), we computed precision, recall, and F1 for each category relative to Coder B's annotations (see Table~\ref{tab:rq3_per_category}  and Figure~\ref{fig:rq3_percategory_f1}).

The results reveal several consistent patterns. \textit{NonKC} was the easiest category to identify in all domains, with F1 ranging from .828 to .954. \textit{Integrate} also showed robust performance across domains, with F1 between .779 and .826, consistent with the within-domain results reported in  Section~\ref{sec:per-class}, where \textit{NonKC} and \textit{Integrate} were likewise the best-performing categories. \textit{Explore} showed moderate yet stable performance, with F1 ranging from .693 to .748, indicating that the classifier captured exploratory behaviours with reasonable consistency. In contrast, \textit{Share} exhibited the greatest cross-domain variability and was clearly the most fragile category. Performance was relatively stronger in the Medicine (F1 = .778) and Programming (F1 = .661) domains, but dropped substantially in Language (F1 = .455) and Music (F1 = .543).

\begin{table}[htbp]
\centering
\caption{Per-category F1 scores by domain (all samples: Random + Augment), computed against coder~B's labels.}
\resizebox{\linewidth}{!}{
\begin{tabular}{lrrrr}
\hline
\textbf{Category} & \textbf{Medicine} & \textbf{Language} & \textbf{Music} & \textbf{Programming} \\
\hline
Explore   & .746 & .733 & .693 & .748 \\
Integrate & .826 & .779 & .809 & .824 \\
Share     & .778 & .455 & .543 & .661 \\
NonKC     & .923 & .891 & .828 & .954 \\
\hline
\end{tabular}}
\label{tab:rq3_per_category}
\end{table}

\begin{figure*}[h]
    \centering
    \includegraphics[width=\linewidth]{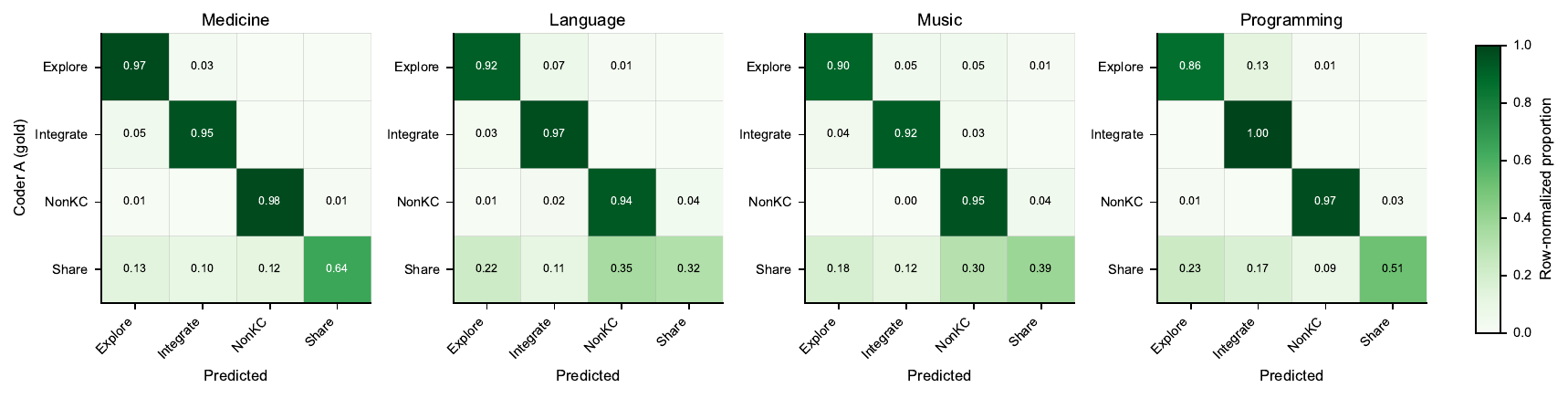}
    \caption{Confusion matrices of model-assigned categories across Medicine, Language, Music, and Programming, compared against coder~B’s labels (all samples: Random + Augment).}
    \label{fig: confusion matrix}
\end{figure*}

To better understand the error patterns behind these scores, we examined the row-normalised confusion matrices in Figure~\ref{fig: confusion matrix}. Across all four domains, most \textit{Explore} and \textit{Integrate} comments lie on the diagonal (typically 90--97\% correct for \textit{Explore} and 92--100\% for \textit{Integrate}), with the main off-diagonal mass corresponding to mutual confusion between these two adjacent stages rather than with \textit{NonKC}. \textit{NonKC} is also clearly separated: 94--98\% of gold \textit{NonKC} comments are correctly classified, and misclassifications into knowledge-construction categories are rare and mostly routed through \textit{Share}. In contrast, errors are concentrated in the \textit{Share} rows. In Medicine and Programming, many \textit{Share} comments are redistributed to \textit{Explore} or \textit{Integrate}, whereas in Language and Music a substantial proportion of \textit{Share} comments are instead predicted as \textit{NonKC} (e.g., 35\% in Language and 30\% in Music). This pattern likely reflects that certain domain-specific behaviours, such as sentence construction in language learning or expressive performance-related actions in music learning, were coded by humans as \textit{Share} but classified by the model as \textit{NonKC} because such behaviours were absent from the training data. It may also be influenced by the fact that \textit{Share} is the category with the lowest overall classification performance, partly because it encompasses a broader and more heterogeneous set of behaviours than the other categories. 

In summary, the results indicate that the classifier can generalise across educational domains, but with limitations. At the aggregate level, the model retained much of its within-domain performance in the Medicine and Programming domains, whereas accuracy decreased substantially in the Language and Music domains. At the category level, the transformer model’s strengths were not confined to a single category, \textit{NonKC}, \textit{Explore}, and \textit{Integrate} all achieved moderate to substantial F1 scores across domains. The primary cross-domain weakness was \textit{Share}, particularly in domains with strong domain-specific discourse features, suggesting a need for domain-sensitive calibration or additional training data.

\section{Discussion}
\label{sec:discussion}

This study developed and validated a framework combining a theoretically grounded codebook with a high-performing transformer classifier to analyse knowledge construction (KC) in non-linear, loosely connected online discourse. Our findings demonstrate that adapting interaction-oriented models like the IAM to comment-level classification is feasible and informative at scale.

The revised four-category IAM scheme preserves core conceptual distinctions \parencite{Gunawardena1997} while addressing limitations in unstructured environments \parencite{DeWever2006,Dubovi2020}. By consolidating higher-level phases into \textit{Integrate} and incorporating a NonKC category \parencite{Nguyen2023}, the scheme trades some granularity for coding reliability, achieving performance comparable to or exceeding earlier adaptations \parencite{Gunawardena2023,Nguyen2023}.

Unlike prior approaches dependent on sequential interactions, our framework detects epistemic moves without requiring thread continuity. Cross-domain evaluation shows that \textit{Integrate} generalises well, while \textit{Share} exhibits greater linguistic variability, indicating that category design influences transferability. This positions the adapted IAM as suitable for large-scale, comment-based environments such as YouTube.

Transformer models, particularly DeBERTa variants, outperformed traditional TF--IDF approaches, aligning with recent NLP research \parencite{Ba2023,Castellanos-Reyes2025,He2021}. Ablation analyses suggest that additional techniques like label smoothing mainly improve stability and calibration rather than accuracy, indicating that well-configured base transformers suffice for many KC tasks. The model’s ability to generalise \textit{Explore} and \textit{Integrate} across domains suggests it captures deeper epistemic functions beyond surface lexical patterns, whereas \textit{Share} remains sensitive to pragmatic variation. These results also connect back to the construct-validity concerns discussed in Section~\ref{sec:model limitations}. The fact that Explore and Integrate generalised more consistently across domains than might be expected from surface lexical patterns indicates that the model is not merely learning topic-specific cues, but is partially capturing deeper epistemic functions. Conversely, the instability of Share aligns with the prediction that heterogeneous, pragmatically variable categories are more vulnerable to lexical approximation. This provides empirical evidence that automated IAM coding can approximate epistemic constructs, but only when those constructs have sufficiently distinct linguistic realisations.

Cross-domain evaluation provides further insight into the strengths and limits of automated KC classification. Similar to findings in domain adaptation research \parencite{peters2019tune,gururangan2020don}, our results show that models trained in one domain can generalise to others, but performance varies depending on the linguistic and pragmatic characteristics of the target domain. In structured and content-heavy domains such as medicine and programming, Explore and Integrate categories remained relatively stable, echoing observations that Transformers are adept at capturing domain-general semantic patterns \parencite{He2021}. However, the reduced performance on Share in language- and music-related channels mirrors prior reports that low-elaboration contributions often rely on subtle pragmatic markers \parencite{Nguyen2023} that differ across disciplinary and community contexts. These patterns indicate that some aspects of knowledge construction are more domain-sensitive than others, and they underline the value of light domain adaptation or selective human review for categories with high linguistic variability.

Overall, this study shows that IAM-based coding can be reliably adapted for large, non-interactive comment streams and that modern transformers can support semi-automated KC analysis at unprecedented scale and performance in informal online learning environments.

\subsection{Implications for research and practice}
\label{sec:discussion_implications}

This study demonstrates a practical approach for adapting the IAM framework to large-scale, nonlinear, noninteractive datasets. By redefining the phases as comment-level categories and incorporating an explicit NonKC label, we preserved meaningful theoretical distinctions while producing a schema that is more feasible for annotation and modeling. This approach enables researchers to conduct reliable content analyses in large, fragmented online learning environments and supports practitioners in efficiently distinguishing learning-related from non-learning contributions.

The results also indicate that transformer-based models can effectively support, rather than replace, human discourse analysis. Although DeBERTa-KC does not reach the level of human–human agreement, its performance is sufficient for generating aggregate indicators and enabling semi-automated workflows, and it clearly surpasses traditional machine-learning baselines. Prediction entropy further provides a useful mechanism for identifying low-confidence outputs that require human review.

Cross-domain analyses suggest that our classifier generalises beyond its training context, particularly in domains with more structured and task-oriented discourse. The model provides reliable signals for NonKC, Explore, and Integrate, whereas Share remains the most sensitive to domain-specific linguistic variation. This behaviour not only highlights the need for light domain calibration or selective human oversight when deploying automated KC classification, but also offers insight into the IAM categories themselves. Categories with clearer conceptual boundaries, such as Explore and Integrate, were consistently recoverable across domains, while Share, operationalised as a broad and heterogeneous category, showed substantially weaker stability. These patterns suggest that future adaptations of the IAM for large and unstructured environments may benefit from refining or subdividing Share to reduce its pragmatic variability.

\subsection{Limitations and directions for future work}
\label{sec:discussion_limitations}

First, our reliability analyses treat comments as independent units and do not compute cluster-adjusted agreement coefficients by video or domain, which may differ somewhat from reliability estimates obtained on fully natural, imbalanced samples. Although we adapted IAM to four broad categories and demonstrated high reliability, the scheme still represents a simplification of the original five-phase model and of the rich spectrum of online epistemic moves. In particular, the combined \textit{Integrate} category conflates activities such as testing, integration, and application. Future work could explore whether more fine-grained subcategories within \textit{Integrate} can be recovered reliably in specific domains or with additional contextual information.

Second, our modeling approach treats each comment as an isolated unit. However, some platforms support threaded interactions, enabling comments to unfold within conversational structures. Incorporating sequence models that account for thread structure or temporal dynamics may better capture how knowledge construction develops over time.

Third, although we examined cross-domain generalisation on four additional YouTube channels, all data in this study come from a single platform and language context. Applying the approach to different platforms and other languages or cultural settings would provide a stronger test of generalisability and might require further adaptation of the coding scheme. 

Fourth, the external validation domains, while diverse, are not fully representative of all educational uses of social media. For instance, we did not include domains with highly contentious or polarised debates, where epistemic and socio-political dimensions are deeply intertwined, nor did we examine settings with strong teacher presence and scaffolding. Investigating how IAM categories manifest in such contexts could refine both the coding scheme and model behaviour.

Finally, we did not explore more advanced forms of model adaptation or fairness-aware training. Domain adaptation techniques (e.g., adversarial alignment, parameter-efficient fine-tuning) might improve performance in challenging domains without requiring extensive re-annotation. Likewise, future work should examine whether model errors are systematically patterned by learner characteristics (where available) or by topic, and consider bias-sensitive evaluation and mitigation strategies.

\section{Conclusion}
\label{sec:conclusion}

This study proposed and validated a framework for analyzing knowledge construction within nonlinear and unstructured text in online learning environments. By adapting four categories of knowledge construction from the Interaction Analysis Model, we developed a codebook suited to these environments, including NonKC, Share, Explore, and Integrate. Building on this codebook, we introduced DeBERTa-KC, an automated classifier based on the DeBERTa-V3-large architecture, which outperformed baseline models and showed stable generalization in external validations. The categories of NonKC, Explore, and Integrate transferred more consistently across contexts, whereas Share appeared more context dependent. Overall, this study provides an empirically validated and scalable approach for identifying knowledge construction in unstructured learner discourse. The findings open new opportunities for systematic and large-scale investigations of learners' cognitive processes in online learning environments and offer meaningful implications for learning analytics and future research.




\printbibliography

\appendix

\section{Appendix A}
\newpage

\begin{table*}[htbp]
\centering
\caption{Coding book for knowledge construction categories (Part 1).}
\begin{tabularx}{\textwidth}{p{1.6cm} |p{2.8cm} |X |X |X}
\toprule
\textbf{Category} & \textbf{Indicator} & \textbf{Definition} & \textbf{Exclusion} & \textbf{Example} \\
\midrule
\multirow{4}{*}{NonKC} 
  & Express emotion 
  & Comments that express emotions or exaggerated reactions without reference to knowledge or reasoning. 
  & If the emotion evaluates or comments on the scientific stance $\rightarrow$ Share / Explore. 
  & ``lol'' \\[0.3em]

  & Socialise or acknowledge others 
  & Greetings, compliments, or support without content-related viewpoints. 
  & If cognitive evaluation of content is present $\rightarrow$ Share / Explore. 
  & ``Love your channel---keep it up!'' \\[0.3em]

  & Produce irrelevant or incoherent content 
  & Symbols, stretched letters, or random fragments lacking interpretable meaning. 
  & If any meaningful knowledge-related content is included $\rightarrow$ Share / Explore. 
  & ``asdjkl\ldots\ldots{} !!'' \\[0.3em]

  & Ask unrelated questions to the learning content 
  & Questions about the creator, platform, or unrelated topics.  
  & If asking about content concepts $\rightarrow$ Share; if challenging $\rightarrow$ Explore. 
  & ``When's your TV show?'' \\[0.6em]
\hline
\multirow{5}{*}{Share} 
  & State an opinion or observation 
  & Personal viewpoint or observation without conflict, reasoning, or synthesis. 
  & If contradicts or refutes $\rightarrow$ Explore; if integrates $\rightarrow$ Integrate. 
  & ``Heat rises!'' \\[0.3em]

  & Express simple agreement 
  & Agreement without justification or elaboration. 
  & If reasons or evidence are added $\rightarrow$ Explore / Integrate. 
  & ``Totally agree!'' \\[0.3em]

  & Provide simple example or information 
  & One additional fact or instance without argumentative intent. 
  & If used to challenge or evaluate $\rightarrow$ Explore; if comparing sources $\rightarrow$ Integrate. 
  & ``The computer already has night-light settings.'' \\[0.3em]

  & Ask for clarification 
  & Questions seeking information but not challenging correctness. 
  & If questioning accuracy or logic $\rightarrow$ Explore. 
  & ``Why is the balloon doing that and then it stayed like that?'' \\[0.3em]

  & Identify a problem or issue 
  & Pointing out an issue or uncertainty about the topic without challenging or evaluating reasoning. 
  & If directly challenges $\rightarrow$ Explore; if integrating solutions $\rightarrow$ Integrate. 
  & ``The image seems too blurry to tell what’s actually happening.'' \\[0.6em]
\bottomrule
\end{tabularx}
\label{tab:coding_book1}
\end{table*}

\begin{table*}[htbp]
\centering
\caption{Coding book for knowledge construction categories (Part 2).}
\begin{tabularx}{\textwidth}{p{1.6cm} |p{2.8cm} |X |X |X}
\toprule
\textbf{Category} & \textbf{Indicator} & \textbf{Definition} & \textbf{Exclusion} & \textbf{Example} \\
\midrule
\multirow{3}{*}{Explore} 
  & Express disagreement 
  & Disagreement or doubt without integrating multiple perspectives. 
  & If synthesising or proposing a new framework $\rightarrow$ Integrate. 
  & ``You got frequency and wavelengths mixed up.'' \\[0.3em]

  & Question or probe another statement 
  & Questions evaluating correctness, logic, or evidence. 
  & If merely seeking information $\rightarrow$ Share. 
  & ``Isn’t platysma just a degenerated version of this?'' \\[0.3em]

  & Defend one’s position with minimal support 
  & Defending a position with one or two reasons but without deeper comparison or integration. 
  & If multiple perspectives are weighed $\rightarrow$ Integrate. 
  & ``Really? I heard it’s Pauli exclusion, not Heisenberg uncertainty.'' \\[0.6em]
\hline
\multirow{4}{*}{Integrate} 
  & Clarify concepts or terms 
  & Refines or corrects conceptual understanding within a broader framework. 
  & If only points out terminology $\rightarrow$ Share. 
  & ``This isn’t about $F = ma$. It’s actually an example of inertia, where the force comes from the wood slowing down.'' \\[0.3em]

  & Propose or synthesise integrative statements 
  & Combines multiple viewpoints or pieces of evidence to propose a new explanation or compromise. 
  & If just taking sides $\rightarrow$ Share. 
  & ``It’s partly about buoyancy and partly about surface tension. The two together explain why it behaves differently in saltwater.'' \\[0.3em]

  & Evaluate statements by comparing perspectives, experiences, or information sources 
  & Compares sources, consequences, or perspectives to draw a judgment. 
  & If only a single piece of evidence is used $\rightarrow$ Explore. 
  & ``Studies say blue light isn’t the main cause, but many people still report less strain with filters, so the effect may be more behavioural than optical.'' \\[0.3em]

  & Apply or reflect on new understanding 
  & Shows changed understanding or connects content to broader implications. 
  & If only noting that something was learned without explicit reflection $\rightarrow$ Share. 
  & ``I always thought going low-carb was the only way to manage blood sugar. But after seeing people compare research showing that increasing fibre can help just as much, I tried adding beans and whole grains instead of cutting carbs completely. What surprised me is that it actually feels easier to stick to, and my energy swings are way smaller now.'' \\
\bottomrule
\end{tabularx}
\label{tab:coding_book2}
\end{table*}


\end{document}